\documentclass[10pt,twocolumn,letterpaper]{article}

\usepackage[pagenumbers]{cvpr} 

\usepackage{url}            
\usepackage{booktabs}       
\usepackage{amsfonts}       
\usepackage{nicefrac}       
\usepackage{microtype}      
\usepackage{xcolor}         
\usepackage{graphicx}
\usepackage{amsmath,dsfont}
\usepackage{adjustbox}
\usepackage{multirow}
\usepackage{mathtools}
\usepackage{amsthm}
\usepackage{bbm}
\usepackage[accsupp]{axessibility}

\definecolor{cvprblue}{rgb}{0.21,0.49,0.74}
\usepackage[pagebackref,breaklinks,colorlinks,allcolors=cvprblue]{hyperref}

\def\confName{CVPR}
\def\confYear{2025}

\title{Rethinking Training for De-biasing Text-to-Image Generation:\\Unlocking the Potential of Stable Diffusion}

\author{Eunji Kim$^{1,}$\footnotemark[1] \qquad Siwon Kim$^{1,}$\footnotemark[1] 
  \qquad Minjun Park$^{2}$
  \qquad Rahim Entezari$^{3,}$\footnotemark[2] 
  \qquad Sungroh Yoon$^{1,2,4,}$\footnotemark[2]\\
$^1$Department of Electrical and Computer Engineering, Seoul National University\\
$^2$Interdisciplinary Program in Artificial Intelligence, Seoul National University\\
$^{3}$ Stability AI \qquad
$^4$AIIS, ASRI, INMC, ISRC, Seoul National University\\
{\tt\small \{kce407, tuslkkk, minjunpark\}@snu.ac.kr, rahim.entezari@stability.ai, sryoon@snu.ac.kr}
}

\begin{document}
\maketitle

\renewcommand{\thefootnote}{\fnsymbol{footnote}}
\footnotetext[1]{Co-first authors}
\footnotetext[2]{Senior authorship with Rahim Entezari and Sungroh Yoon (corresponding author: Sungroh Yoon)}
\renewcommand{\thefootnote}{\arabic{footnote}}

\begin{abstract}
Recent advancements in text-to-image models, such as Stable Diffusion, show significant demographic biases. Existing de-biasing techniques rely heavily on additional training, which imposes high computational costs and risks of compromising core image generation functionality. This hinders them from being widely adopted to real-world applications. In this paper, we explore Stable Diffusion's overlooked potential to reduce bias without requiring additional training. Through our analysis, we uncover that initial noises associated with minority attributes form ``minority regions'' rather than scattered. We view these ``minority regions'' as opportunities in SD to reduce bias. To unlock the potential, we propose a novel de-biasing method called `weak guidance,' carefully designed to guide a random noise to the minority regions without compromising semantic integrity. Through analysis and experiments on various versions of SD, we demonstrate that our proposed approach effectively reduces bias without additional training, achieving both efficiency and preservation of core image generation functionality.
\end{abstract}    
\section{Introduction}
\label{sec:intro}

With their ability to generate high-quality images, diffusion-based text-to-image (T2I) models are gaining widespread use across various platforms~\cite{esser2024scaling, podell2023sdxl, rombach2022high, ramesh2022hierarchical, team2023gemini}. However, concerns have emerged about biases in these models, particularly the severe underrepresentation of minority groups in terms of gender, race, or other demographics, diverging from real-world distribution. In some cases, this even has led to some service backlash~\cite{barroso2024racial, cho2023dall, bird2023typology}. Even the recently open-sourced model Stable Diffusion 3 (SD3)~\cite{esser2024scaling} produces biased outputs, \textit{e.g.}, generating nearly 100\% female CEO, showing that these issues persist.

Several approaches have been explored to mitigate these biases, primarily through additional training. Previous efforts have involved fine-tuning all or parts of the model parameters or using techniques like prompt tuning for more efficient learning~\cite{kim2023stereotyping,shen2024finetuning,gandikota2024unified,orgad2023editing}. However, these approaches have substantial drawbacks, including high computational costs and the risk of compromising the core image generation capabilities of T2I models. For most companies deploying T2I models, maintaining high-quality image generation remains the top priority. Consequently, the widespread adoption of existing de-biasing techniques in commercial T2I services like SD3 has been hindered, as they can impair both model performance and scalability.

\begin{figure}
  \centering
    \includegraphics[width=0.88\linewidth]{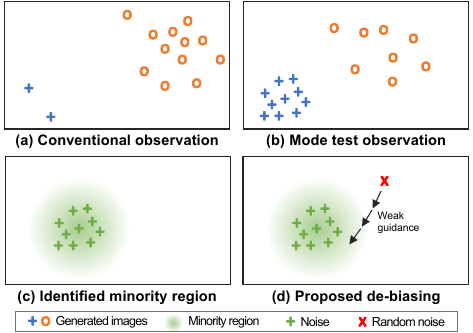}
    \vspace{-0.5em}
    \caption{
    Conceptual illustration of our contributions. \textcolor{blue}{\textbf{+}} and \textcolor{orange}{\textbf{o}} denotes observed minority and majority images. \textcolor{green}{+} denotes noises associated with a minor attribute. Mode test (\Cref{sec:mode_test}) identifies the existence of minority regions (b, c) and the proposed method (\Cref{sec:method}) guides an initial noise to the minority regions (d).
    }
    \label{fig:thumbnail2}
\end{figure}

Is training truly essential to de-bias T2I models? Specifically, can any aspects of the pre-trained Stable Diffusion (SD) model itself be leveraged to mitigate bias? In this study, we explore the existence of ``minority regions'' by using a newly introduced ``mode test.'' We define a ``minority region'' as the area in the noise space whose significant portion is generated to minority images (\textit{e.g.}, female CEO) when conditioned with an attribute-neutral prompt (\textit{e.g.}, ``\texttt{A photo of a CEO}"). Our analysis in \Cref{sec:analyses} reveals that the noises associated with these minor attributes form regions rather than being sparse or randomly scattered, more prevalently than previously observed. This is conceptually illustrated in \Cref{fig:thumbnail2}. These findings suggest that, if we can effectively harness those regions, we can reduce bias without requiring additional training.

\begin{figure*}
  \centering
  \includegraphics[width=0.95\textwidth]{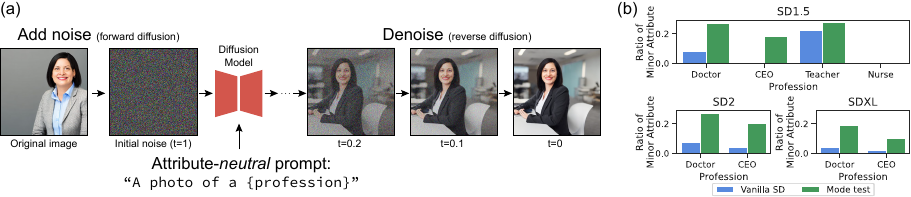}
  \vspace{-0.5em}
  \caption{(a) Conceptual illustration of our mode test. The encoding and decoding steps of the latent diffusion model are omitted from the figure. Noise is added to minor attribute images, followed by reverse diffusion process with an attribute-neutral prompt. (b) Change in the minor attribute ratio with the mode test~(\Cref{sec:mode_test}).}
  \label{fig:modetest}
\end{figure*}

The newly enlightened minority regions inspires the concept of navigating random initial noises toward these regions. To begin, we analyze whether weakening the original guidance can serve as an implicit guidance. Specifically, given that the typical SD employs textual guidance, we test two schemes as follows: 1) reducing the strength of the textual guidance, and 2) adding noise to the textual guidance to perturb its direction. Experimental results demonstrate that both effectively reduce bias, yet they also compromise image-text alignment, underscoring the need for a more refined guidance mechanism. The following analysis reveals that explicitly directing perturbations towards minority regions offers a promising alternative where it impacts only the attribute without compromising image-text alignment. 

Building on these analyses, we propose a novel de-biasing method that adopts more direct guidance to the minority regions. Specifically, attribute directions are randomly selected and added to the textual condition. To preserve the semantics of the original text condition, the addition is conducted in a \textit{weak} manner; the attribute direction is not added to every dimension, but rather to a less meaningful part of the text condition embedding. Our experiments demonstrate that our design choices preserve the image generation capability of SD while ensuring strong de-biasing performance.

Final experimental results demonstrate the effectiveness of our method in mitigating gender and racial bias in various professions while preserving the aesthetic quality and prompt alignment of images generated by SD with minimal impact on overall output. We test our method across multiple SD versions, including SD1.5, SD2, SDXL, and SD3 developed by different entities, distinguishing our work from previous works that were limited to only a single version of SD. This highlights the generalizability of our analysis and the proposed approach.

Our main contributions can be summarized as follows:
\begin{itemize}
\setlength{\itemsep}{2pt}

\item We propose for the first time a mode test to closely examine minority regions, previously perceived as sparse and minuscule, and explore their potential for enabling training-free de-biasing by steering initial noise toward these regions through text guidance adjustment.

\item Motivated by our findings, we propose a novel de-biasing method based on weak text guidance that redirects random noise to generate minority images.

\item We experimentally demonstrate that our method reduces bias while preserving the image generation capability of SD.
Results across multiple versions of SD further assure the broad applicability and effectiveness of our method.
\end{itemize}

\section{Related Works}
\label{sec:relatedworks}

\paragraph{Analysis on bias in text-to-image generation}
Recent studies have analyzed biases in T2I models. \citet{luccioni2023stable} and \citet{perera2023analyzing} brought attention to how these models often produce images that reflect societal stereotypes. \citet{bianchi2023easily} and \citet{seshadri2023bias} demonstrated that T2I models can amplify stereotypes, leading to skewed representations in generated images. While these works provide valuable insights into the existence of bias, they largely focus on analyzing outputs rather than deeply investigating the potential of harnessing SD's property to reduce such biases.

\paragraph{De-biasing with additional training} 
Training-based de-biasing approaches aim to introduce fairness to SD by using additional resources. However, fully fine-tuning large T2I models is prohibitively costly. Recent approaches have adopted parameter-efficient fine-tuning techniques, such as prefix tuning~\cite{kim2023stereotyping}, text embedding projection weight~\cite{chuang2023debiasing}, or low-rank adaptation \cite{shen2024finetuning}. Other efforts have focused on modifying the cross-attention layer in the UNet of SD~\cite{gandikota2024unified, orgad2023editing}. Some studies also explore fine-tuning h-space vectors, located in the UNet’s bottleneck layer, known to encapsulate rich semantic information~\cite{li2024self, parihar2024balancing}. However, there has been little examination of whether additional training is essential.

\paragraph{De-biasing without additional training}
There are very few existing de-biasing techniques that bypass additional training entirely, typically opting to modify text prompts instead.
ENTIGEN~\cite{bansal2022well} inserts ethics-oriented phrases to prompts. However, manipulating the prompt explicitly risks distorting the semantic integrity of images generated with generic prompts that do not induce bias.
FairDiffusion~\cite{friedrich2023fair}, the work most similar to ours, modifies the diffusion direction by employing SEGA~\cite{brack2023sega}, a concept editing technique. However, as discussed in \Cref{sec:exp}, FairDiffusion tends to compromise image-text alignment, especially when a major attribute is explicitly mentioned in the text prompt or when prompt is devoid of potential stereotypes. 
In contrast, our method is carefully designed to ensure the text-image alignment after de-biasing.

\section{Exploring the Potential for Training-free De-biasing}
\label{sec:analyses}

In this section, we delve into the minority regions as the potential for training-free de-biasing. First, we propose a mode test in \Cref{sec:mode_test}, which examines the existence and range of the minority regions.  In \Cref{sec:text_condition}, we explore text condition as a means of guiding initial noise to the enlightened minority regions.

The primary analysis centers on binary gender bias (male and female) across four professions (doctor, CEO, nurse, and teacher). To classify gender in the generated images, we apply the CLIP zero-shot classifier\footnote{\url{https://huggingface.co/openai/clip-vit-base-patch32}} with prompts such as ``\texttt{A photo of a male/female}". For racial bias testing, we follow \citet{chuang2023debiasing} and use text prompts like ``\texttt{A photo of a White person/Black person/Asian/Indian/Latino}". The attribute that appears most frequently in generated images is identified as the major attribute, with other attributes labeled as minor. Further details are described in the Appendix.

\subsection{Exploring Minority Regions: Mode Test}\label{sec:mode_test}
To motivate the mode test, we first clarify its necessity. The conventional bias assessment has been conducted by generating a finite number of images, typically a few hundred to a few thousand, and counting the number of minority images, \textit{e.g.}, 30 female CEO images out of 1,000 CEO images (SD1.5). This approach led to the perception that the minority regions are sparse and minuscule (\Cref{fig:thumbnail2} (a)) suggesting that additional training is essential. If it is possible to generate images infinitely, all existing minority regions will be fully uncovered. However, since this is impractical, the mode test helps to explore the existence of minority regions without requiring infinite generation.

Mode test works as follows; we purposefully generate minority images (\textit{e.g.}, female CEO) by using prompts that specify minor attributes (``\texttt{A photo of a female CEO}").
Then we add standard Gaussian noise to those images to convert the images back to the initial noise space.
This can be thought of as simulating a forward diffusion process similar to SDEdit~\cite{meng2022sdedit}. 
Then the noises are denoised by given attribute-\textit{neutral} prompts (\textit{e.g.}, ``\texttt{A photo of a CEO}"). 
The overall process of the mode test is illustrated in \Cref{fig:modetest}(a). 
Because the added noise is sampled from a standard Gaussian distribution, the resulting noise points are likely to be located near the regions corresponding to the original images.
Therefore, if minor attributes reappear in the images generated from \textit{attribute-neutral} prompts, it suggests that the noises associated with minority attributes are not randomly scattered but instead clustered closely, forming a defined region. 
This insight reveals the structured nature of these regions and their potential for training-free de-biasing.

\begin{figure}[t]
\centering
\includegraphics[width=0.85\linewidth]{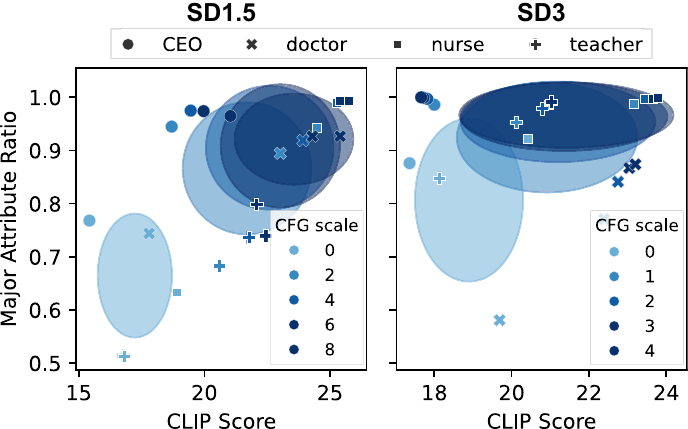}
 \vspace{-0.5em}
\caption{Impact of CFG: 
A higher CFG scale raises both the ratio of major attributes and the CLIP score~(\Cref{sec:cfg}).}
  \label{fig:cfg_scale_bias}
\end{figure}
\begin{figure*}[t]
  \centering
  \includegraphics[width=\linewidth]{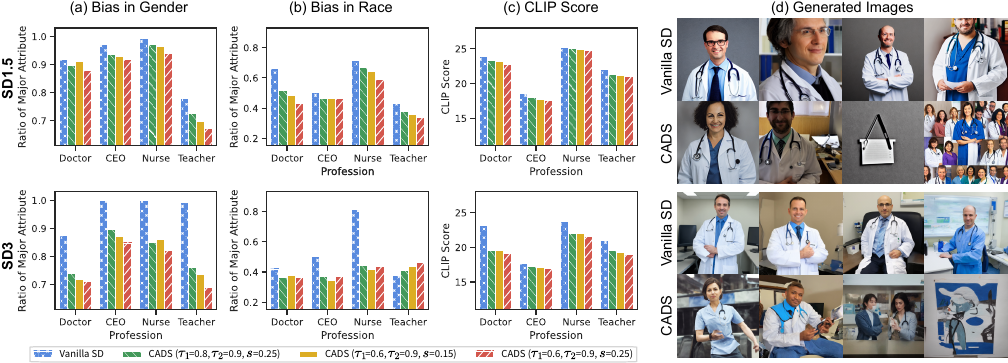}
  \caption{Impact of CADS on bias and CLIP score: Increasing noise injection in the text condition (higher $s$ and lower $\tau_1$) effectively reduces bias (a, b) but also lowers the CLIP score (c). (d) Generated samples from vanilla SD and CADS ($\tau_1=0.6, \tau_2=0.9, s=0.25$) using the prompt `\texttt{a photo of a doctor}". CADS improves diversity in gender and race representation, though occasional misalignment with the prompt is observed (\Cref{sec:analyses_cads}).}
  \label{fig:cads_sd1.5}
\end{figure*}

\Cref{fig:modetest} (b) presents a comparison of the minor attribute ratio between images generated by vanilla SD and those produced through the mode test, with 1,000 images evaluated for each profession.
The results show that the ratio of minor attributes increases in mode test generation compared to the vanilla SD.
These suggest the existence of minority regions, \textit{i.e.,} the initial noises are clustered rather than sparse and randomly dispersed, motivating us to harness those regions.

\subsection{Exploring Guidance to the Minority Region}\label{sec:text_condition}

In this section, we propose that the text condition can steer initial noise toward the minority regions. 
To test this, we perform two experiments: (1) reducing the classifier-free guidance scale (\Cref{sec:cfg}) and (2) introducing noise into the text conditions (\Cref{sec:analyses_cads}).

\subsubsection{Impact of Classifier Free Guidance}\label{sec:cfg}
The Classifier-Free Guidance (CFG) \cite{ho2021classifier} guides an initial noise to the final image that depicts the semantics specified in the text condition.
Specifically, with CFG, the predicted noise $\Tilde{\epsilon}_\theta$ can be written as $\Tilde{\epsilon}_\theta(\boldsymbol{z}, \boldsymbol{c}) = (1+\alpha)\cdot \epsilon_\theta(\boldsymbol{z}, \boldsymbol{c}) - \alpha \cdot \epsilon_\theta(\boldsymbol{z})$,
where $\boldsymbol{z}$ represents the unconditional embedding, $\boldsymbol{c}$ is the conditional embedding derived from the text prompt, and $\alpha$ is the CFG scale.
It is well-established that a larger value of $\alpha$, representing stronger guidance, results in higher coherence between the generated image and the text prompt, but this comes at the expense of sample diversity~\cite{ho2021classifier}.

In this study, we examine how bias changes as the CFG scale varies, generating a total of 5,000 images for each profession (five CFG scales $\times$ 1,000 generations per profession). 
\Cref{fig:cfg_scale_bias} presents the ratio of the major attribute (y-axis) against the CLIP score (x-axis), with color intensity indicating the magnitude of the CFG scale. 
As the CFG scale decreases (represented by lighter colors), the ratio of the major attribute also declines. 
These findings support our hypothesis that weakening the text condition can reduce bias, though it comes at the cost of decreased alignment between the generated images and the text prompts.

\subsubsection{Noisy Text Condition}\label{sec:analyses_cads}
We explore an alternative method for weakening text conditions by injecting noise, inspired by Condition-Annealed Sampling (CADS)~\cite{sadat2024cads}. CADS introduces noise to text conditions to diversify the composition of generated images.
CADS perturbs a given text condition $c$ to $\hat{c}$ as follows:
\begin{equation}
\label{eq:cads}
\begin{aligned}
&\hat{\boldsymbol{c}} = \sqrt{\gamma(t)}\boldsymbol{c} + s\sqrt{1-\gamma(t)}\boldsymbol{n}, \\
&\text{where}~\gamma(t)=
\begin{cases}
1 & 0 \leq t \leq \tau_1, \\
\frac{\tau_2-t}{\tau_2-\tau_1}~~& \tau_1 < t < \tau_2, \\
0 & \tau_2 \leq t \leq 1.
\end{cases}
\end{aligned}
\end{equation}
Here, $s$ determines the noise scale, $\gamma(t)$ represents the annealed coefficient dependent on time step $t$, and $\boldsymbol{n}\sim\mathcal{N}(0, I)$ denotes Gaussian noise.
As diffusion models generate images through a reverse process from $t=1$ to $t=0$, noise perturbations are applied to a text condition in earlier steps.
$\hat{\boldsymbol{c}}$ is then normalized to have the same mean and standard deviation as $\boldsymbol{c}$.
To study the effect of noise injection to text conditions on diversifying attributes, we conduct experiments addressing gender and racial bias.
Building on CADS default settings, we extend our analysis to explore the impact of noise intensity and injection duration.

The results in \Cref{fig:cads_sd1.5}(a,b) demonstrate the effect of CADS on reducing the ratio of major attributes (y-axis), indicating bias mitigation. Across all variations in intensity and duration, the major attribute ratio decreases compared to vanilla SD (blue) for both gender and racial bias. Strong perturbations-achieved by increasing $s$ from 0.15 (yellow) to 0.25 (red) or decreasing $\tau_1$ from 0.8 (green) to 0.6 (red)-further enhance bias reduction.
These observations also show that weakening text conditions helps mitigate bias.
\Cref{fig:cads_sd1.5}(d) compares the images generated with vanilla SD and CADS.
While CADS-generated images display diverse gender and race attributes, the alignment between prompt and generated images degrades, as shown in 3$^{rd}$ and 4$^{th}$ columns of the generated samples.
This decline in alignment is further reflected in \Cref{fig:cads_sd1.5}(c), which shows a decrease in CLIP scores with CADS.
Similar trends are observed for SD2 and SDXL (please refer to Appendix).
These results indicate that, although perturbing text conditions can effectively reduce bias, it is important to carefully design the guidance to preserve SD's image quality and maintain prompt alignment. 

\begin{figure}[t]
\centering
\includegraphics[width=0.99\linewidth]{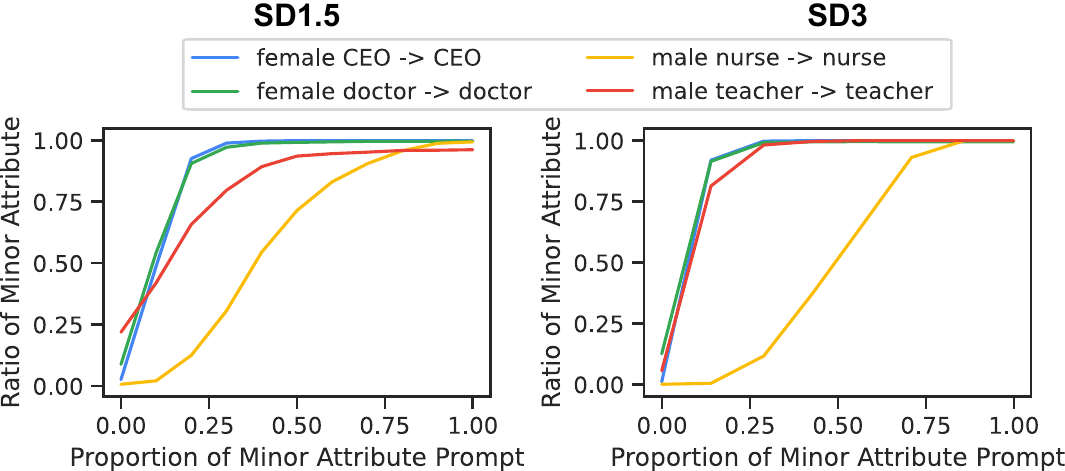}
 \vspace{-0.5em}
  \caption{Ratio of the minor attribute. The x-axis represents the fraction of diffusion steps guided by minor attribute-specific prompts~(\Cref{sec:analysis_minor_attr}).
  }
  \label{fig:twotext_sd15}
\end{figure}

\subsection{Attribute Guidance}\label{sec:analysis_minor_attr}
We examine whether introducing a minor attribute-specified prompt during partial diffusion steps enhances bias reduction.
Here’s the approach: for a neutral prompt, the text condition in the early diffusion stages is replaced with a prompt focused on a minor attribute. The neutral prompt is then provided for the remaining steps.

\Cref{fig:twotext_sd15} shows how varying the proportion of minor attribute-specified prompts (x-axis) influences the minor attribute ratio (y-axis), with different lines representing various professions.
When minor attribute prompts are not used at all (proportion = 0), the results are heavily biased. On the other hand, using only minor attribute prompts (proportion = 1) yields a minor attribute ratio near 1.
The gradual increase in the proportion of minor prompts clearly increases the minor attribute ratio, demonstrating that guidance during early diffusion steps is an effective strategy for addressing bias.
This inspires our proposed de-biasing method in the next section which controls text condition perturbation through minor attribute guidance.

\section{Unlocking the Potential for Fairness in SD}
\label{sec:method}

Building on our previous observations, we propose a straightforward yet effective method to reduce the influence of text prompts that tend to favor major attributes by perturbing the diffusion direction.
This approach balances applicability with SD’s image generation capability. Specifically, it focuses on two goals:
\textbf{1) Efficiency:} The method should not require high costs for training or inference.
\textbf{2) Versatility:} It should address bias only when needed, without hindering general-purpose image generation, such as responding accurately to prompts like ``\texttt{A photo of a car}".
In the following section, we introduce our \textit{weak} conditioning scheme designed to satisfy both objectives.

\subsection{Method Design}\label{sec:mask_embedding}

T2I models, including SD, rely on text encoders such as CLIP~\cite{radford2021learning}, which support arithmetic operations within their embedding space. These operations allow for the addition and subtraction of conceptual directions~\cite{couairon2022embedding}. 
We define the attribute embedding direction $\boldsymbol{a}_k$ for an attribute $k$ as $\phi(k)-\phi(``")$, where $\phi(\cdot)$ represents the text encoder's encoding function, and ``" denotes an empty text.
Here, $k$ is the text representing an attribute, such as ``\texttt{female}" and ``\texttt{male}" for gender bias.
By adding $a_k$ to a given text condition embedding $c$, we obtain a modified embedding $\boldsymbol{c} + \boldsymbol{a}_k$).
This guides the diffusion process toward $k$ without requiring additional training, ensuring computational \textbf{efficiency}.

\begin{figure}[t]
\centering
    \includegraphics[width=\linewidth]{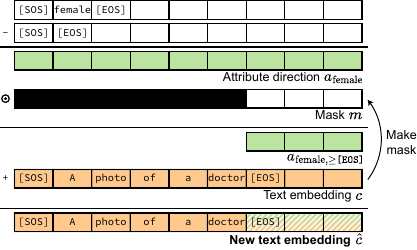}
  \caption{Our weak text embedding creation with attribute embedding. This approach guides the diffusion direction effectively while preserving the original text's semantics~(\Cref{sec:mask_embedding}).}
  \label{fig:weak_attr_embedding}
\end{figure}

To ensure \textbf{versatility}, we redesign the perturbation to be \textit{weak}, ensuring that it maintains the semantic integrity of the input.
In T2I models, the text encoder processes fixed-length inputs by appending an end-of-sequence (\texttt{[EOS]}) token to mark the end of the meaningful text, with padding added to fill the remaining length.
Instead of adding $\boldsymbol{a}_k$ to all positions of $\boldsymbol{c}$, we add a portion of $\boldsymbol{a}_k$ only to the positions starting from the \texttt{[EOS]} token up to the maximum length of $\boldsymbol{c}$.
This selective addition directs the diffusion process toward the desired attribute while preserving the original semantics of the prompt up to the \texttt{[EOS]} token. The benefits of this approach are explored further in \Cref{sec:justification}.
In formal terms, our new text embedding is expressed as $\hat{\boldsymbol{c}} = \boldsymbol{c} + \boldsymbol{m}\odot \boldsymbol{a}_k$, 
where $m_i=\mathds{1}(i \geq \texttt{[EOS]})$ with 
$m$ being a mask whose value is 1 for the positions starting from the \texttt{[EOS]} token onward, and $0$ elsewhere.
\Cref{fig:weak_attr_embedding} illustrates our text embedding creation process.

Given a prompt, we sample a target attribute from a uniform distribution over a pre-defined attribute set. After obtaining the weak attribute guidance, we use $\hat{c}$ exclusively for the initial $\tau$ denoising steps to establish the target attribute's influence. Following this, $c$ and $\hat{c}$ are alternated throughout the remaining denoising steps. This phased approach helps ensure the effective incorporation of the target attribute while preserving the original text's semantics.

While our bias mitigation method appears straightforward, it meets the key goals of being both computationally practical and versatile.
Experimental results in \Cref{sec:exp} will further highlight the effectiveness of this approach.

\subsection{Investigating the Design Choices}\label{sec:justification}
\begin{figure}[t]
  \centering
  \includegraphics[width=0.95\linewidth]{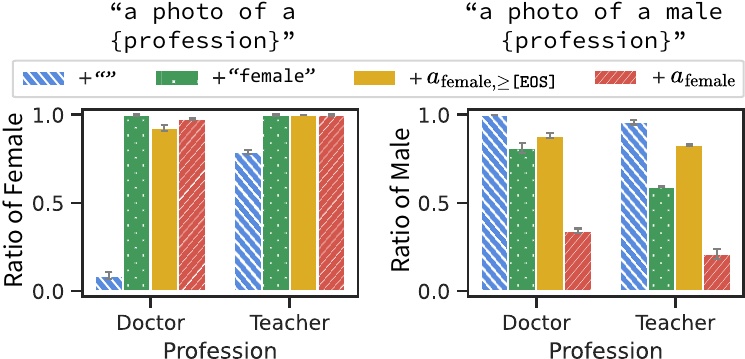}
  \vspace{-0.5em}
  \caption{Change in ratio of attributes with gender-neutral (left) gender-specified (right) prompts. The y-axis represents the ratio of the targeted attribute with guidance. Higher values indicate better alignment with the intended outcome (\Cref{sec:justification}).}
  \label{fig:embedding_arithmetic_sd15}
\end{figure}

In this section, we explain the rationale behind our design choices and demonstrate their effectiveness in reducing bias, promoting fairness, and preserving versatility compared to alternative methods.
\Cref{fig:embedding_arithmetic_sd15} illustrates the impact of different design choices on attribute ratios. The left plots show the influence of guidance with neutral prompts, while the right plots depict the effects with attribute-specific prompts. Our goal is to achieve weak guidance—a subtle adjustment that encourages the generation of specific attributes (as seen in the high ratios in the left plots) without overriding explicit instructions in the text prompt. For example, if the prompt specifies ``male," the guidance toward ``female" should not interfere (indicated by high ratios in the right plots).

\paragraph{Specifying in the prompt vs. adding direction}
Our method adds attribute embedding \textit{direction} rather than explicitly specifying the attribute in the prompt. Inserting an attribute into a text prompt directly requires parsing the sentence to determine the appropriate placement for the attribute. This process is inefficient and may result in grammatically incorrect sentences, compromising image generation. By contrast, with the adding scheme, there is no need to parse or decide where to insert the attribute, making it more efficient and reliable. Simply appending the attribute text to the prompt (green in the left plots of \Cref{fig:embedding_arithmetic_sd15}) can guide the generation toward the specified attribute. However, this approach compromises versatility and the ability to prioritize the original prompt (the right plots).

\paragraph{Adding the attribute direction at every position vs. upon the \texttt{[EOS]} token}
Adding the attribute direction upon the \texttt{[EOS]} token (yellow) reliably generates the targeted attribute.
This indicates that the embedding upon \texttt{[EOS]} carries sufficient information to guide the generation toward the target attribute.
While adding an attribute embedding at every position (red) also increases the ratio of attributes, it generates mostly female images even with ``male" stated in the prompt, indicating excessive guidance. 
In contrast, adding the direction upon \texttt{[EOS]} generates mostly male images, preserving the prompt's intent. 
This demonstrates the effectiveness of weak perturbation in maintaining the original semantics of a text prompt.

\begin{figure}[t]
\centering
\includegraphics[width=0.98\linewidth]{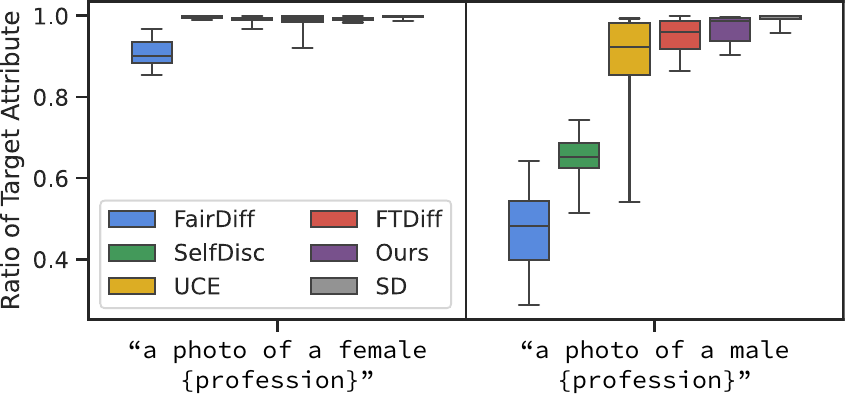}
\vspace{-0.5em}
\caption{Ratio of gender in generations when it is specified in a given text prompt.
If a certain attribute is explicitly specified in the text prompt (\textit{e.g.}, ``\texttt{a photo of a \textit{male} doctor}"), the attribute should be present in generated images. Higher is better~(\Cref{sec:alignment}).
}
\label{fig:result_explicit_attr}
\end{figure}

\begin{table}[t]
\centering
  \begin{tabular}{cc|cc}
    \toprule
    \textbf{Model} & \textbf{Method} & \textbf{CLIP Score} ($\uparrow$) & \textbf{CMMD} ($\downarrow$) \\
    \midrule
    \multirow{5}{*}{SD1.5} &Vanilla SD & \underline{26.42} & \underline{0.532}\\ 
    &FairDiff & {26.03} & 0.586\\
    &UCE &18.10 &1.240\\
    &FTDiff &25.61 &0.783\\
    &SelfDisc &24.97 &0.900 \\
    &Ours & \textbf{26.56} & \textbf{0.509}\\

    \midrule
    \multirow{3}{*}{SD2} &Vanilla SD &  \textbf{26.55} & \textbf{0.524} \\ 
    &FairDiff & 25.91 & 0.602 \\
    &Ours & \underline{26.29} & \underline{0.549}\\
    \midrule
    \multirow{2}{*}{SDXL} &Vanilla SD & 26.51 & 0.794\\
    & Ours & \textbf{26.58} & \textbf{0.757}\\
    \midrule
    \multirow{2}{*}{SD3} &Vanilla SD & \textbf{26.54} & \textbf{0.646}\\
    & Ours &  26.48 & 0.664\\
    \bottomrule
  \end{tabular}
  \caption{Comparison of image fidelity and image-text alignment on MS COCO 2014-30k. Images are generated as $512\times512$ using SD1.5 and SD2 and $1024\times1024$ using SDXL and SD3, respectively~(\Cref{sec:alignment}).}\label{table:fid}
\end{table}

\subsection{Experimental Comparison with Debiasing Methods}
\label{sec:exp}
\paragraph{Baselines}
We compare the proposed method with four baseline methods: FairDiffusion (FairDiff)~\cite{friedrich2023fair},
Unified Concept Editing (UCE)~\cite{gandikota2024unified}, Fine-Tuning Diffusion (FTDiff)~\cite{shen2024finetuning}, and Self-Discovering latent direction (SelfDisc)~\cite{li2024self}. 
FairDiffusion serves as our primary comparison target since it is the only method that does not require additional training.

For all baselines, we use the official code and checkpoints if available. Otherwise, we train from scratch following the authors’ instructions. For additional context, we also report gender statistics of LAION-5B~\cite{schuhmann2022laion}, which comprises the training data for SD~\cite{seshadri2023bias}. Further details are in the Appendix.

\paragraph{Evaluation} 
We evaluate bias using CLIP zero-shot classifier, as described in \Cref{sec:analyses}.
Image-text alignment is measured with the CLIP score~\cite{hessel-etal-2021-clipscore} using the CLIP-ViT-L/14 model\footnote{\url{https://huggingface.co/openai/clip-vit-large-patch14}}, while image fidelity is evaluated with the CMMD score~\cite{jayasumana2024rethinking}. Both metrics are computed on 30,000 images from the MS-COCO-2014~\cite{lin2014microsoft} validation set.

\paragraph{Prompt configurations for debiasing}
We primarily focus on professions where the bias in generated images is more pronounced than in the training data. 
Based on analyses of gender ratios within the training dataset~\cite{seshadri2023bias}, we select eight professions where the difference in the ratio of minor attributes between the training dataset and images generated with SD1.5 exceeds or approaches 10\%p.
These professions span both male-dominated fields (CEO, doctor, pilot, technician) and female-dominated fields (fashion designer, librarian, teacher, nurse). Unless specified otherwise, all images are generated using the template ``\texttt{A photo of a/an {profession}}" to evaluate debiasing performance.

\begin{figure}
  \centering
  \includegraphics[width=\linewidth]{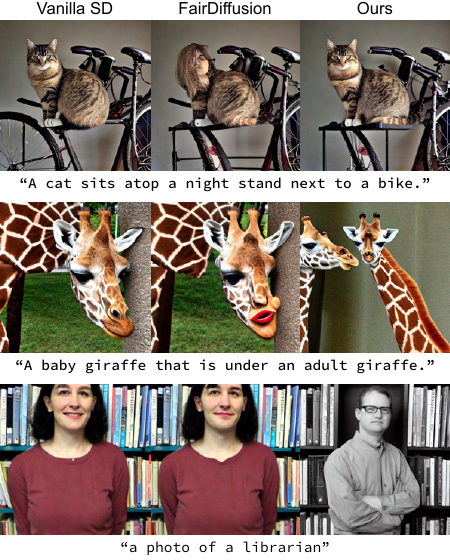}
 \vspace{-2em}
  \caption{Comparison of generations with vanilla SD, FairDiffusion, and our method. Images are generated with captions from COCO-30k (first and second) and a prompt for profession (third) using SD1.5. The third image generated from FairDiffusion (middle) present physical characteristics of both women and men simultaneously~(\Cref{sec:alignment,sec:debiasing}).
 }
  \label{fig:compare_example}
\end{figure}

\subsubsection{Image-Text Alignment and Image Fidelity}\label{sec:alignment}
Our method is designed to uphold the image generation quality of SD while actively pursuing fairness, ensuring that both image-text alignment and image fidelity are preserved.
Otherwise, it may result in significant reliability challenges when being deployed, such as depicting racial demographics inaccurately \cite{GeminiGoogle}.
We have established two fundamental requirements for our quality assurance efforts. The first requirement mandates that if an attribute is explicitly defined in the text prompt (\textit{e.g}, ``\texttt{a photo of a \textit{male} doctor}"), it must be present in the generated images.
The second requirement dictates that our de-biasing technique should not adversely affect the generation of other images.
For instance, a prompt without any potential stereotypes, such as ``\texttt{a photo of a car}", should still yield a high-quality (CMMD score) and accurate car image (CLIP score).

\setlength{\tabcolsep}{3.5pt}
\begin{table*}
  \centering
  \adjustbox{max width=\textwidth}{
  \begin{tabular}{cc|ccccccc}
    \toprule
        Minor attribute & Profession & Training Data~\cite{seshadri2023bias} 
    & Vanilla SD
    & FairDiff~\cite{friedrich2023fair}
    & UCE~\cite{gandikota2024unified}$^\dagger$ 
    & FTDiff~\cite{shen2024finetuning}
    & SelfDisc~\cite{li2024self}$^\dagger$
    & Ours\\
    \midrule
    
    \multirow{4}{*}{\rotatebox[origin=c]{0}{Female}} & CEO & 0.150 & 0.030 & \textbf{0.452 }&0.027 &0.190 &\underline{0.445}	   & 0.389\\
    & Doctor & 0.408 & 0.081 & \textbf{0.502} &0.049 &0.198 &\textbf{0.502} & 0.334\\
    & Pilot & 0.304 & 0.150 & 0.739 &0.244 &0.260 &\textbf{0.568}	 &  \underline{0.408} \\
    & Technician & 0.312 & 0.007  & \textbf{0.553} & 0.005 &0.168 & \underline{0.347} & 0.164 \\
   \midrule
   \multirow{4}{*}{\rotatebox[origin=c]{0}{Male}} & Fashion designer & 0.240 & 0.078 & \underline{0.333} & 0.018 &0.167 &0.067 &\textbf{0.451} \\
    & Librarian  & 0.256 & 0.194 & 0.300 &0.297 &\textbf{0.538} &0.174	 &  \underline{0.421} \\
    & Teacher & 0.370 & 0.222 & 0.205& 0.155 &\underline{0.231} &0.081 & \textbf{0.492} \\
    & Nurse & 0.112 &  0.007 & \underline{0.162} &0.003 &\textbf{0.208} &0.004	 & {0.039} \\
    \midrule
    &Avg. $\Delta$ ($\downarrow$) &-&0.403&\textbf{0.167}&0.400&0.264&0.244&\textbf{0.167}\\
    \bottomrule
  \end{tabular}}
  \caption{Ratio of minor attributes within 1,000 images generated with SD1.5, where a value closer to 0.5 is preferred.
  $\dagger$ indicates results reproduced through training using the official code. Avg. $\Delta$ represents the average absolute difference from the target ratio of 0.5 across all professions~(\Cref{sec:debiasing}).}
  \label{tab:debias_perf_sd15}
\end{table*}

\begin{table}[!htp]\centering
\begin{tabular}{lrrrr}\toprule
&SD2 &SDXL &SD3 \\
\midrule
Vanilla SD &0.444 &0.363 &0.481 \\
Ours &0.362 &0.242 &0.349 \\
\bottomrule
\end{tabular}
\caption{De-biasing results of our method across various versions of SD, with average $\Delta$~(explained in \Cref{tab:debias_perf_sd15}) values reported (lower values indicate better performance)~(\Cref{sec:debiasing}).}\label{tab:debias_perf_general}
\end{table}

\paragraph{Specified attribute}
To test our first requirement, we generate 300 images with text prompts that explicitly specify gender.
\Cref{fig:result_explicit_attr} shows the prompt templates and results for eight professions generated using our method, compared with four baselines and the vanilla SD.
The detailed numerical results are provided in Appendix.
The left plot illustrates the female ratio from female-specified prompts, while the right plot reflects the male-specified prompts. Since gender is explicitly defined in each prompt, \textit{the target ratio is 1}.
The vanilla SD1.5 (grey) achieves a ratio close to 1 in both scenarios.
Our method (purple) also reaches a ratio near 1, indicating that the weak perturbation approach does not disrupt the generation of the specified attribute. Conversely, other methods do not exclusively generate the specified gender. Notably, when using the male-specified prompt (right), FairDiff (blue) shows an almost balanced ratio, indicating that it almost ignores the gender specified in the prompt.

\paragraph{General image generation}
For the second requirement, we generate images using captions from the COCO-30k dataset and evaluate their quality using CLIP and CMMD scores, as presented in \Cref{table:fid}.
Our method achieves scores comparable to the vanilla SD, demonstrating that it maintains both image-text alignment and image fidelity without compromise. In contrast, other methods show decreased CLIP scores and increased CMMD scores, suggesting a notable decline in image generation quality for general text prompts. This issue is particularly pronounced in training-based approaches, which exhibit greater compromises in performance.
Qualitative comparison in \Cref{fig:compare_example} further highlights the superiority of our method over FairDiffusion, our closest competitor.
The images generated by FairDiffusion (second column) frequently include unintended artifacts, such as a hybrid of a human face and a cat (first row) or a giraffe with human-like features (second row), even when the text prompts contain no reference to humans.
In contrast, our method consistently generates coherent and artifact-free images, showcasing its ability to achieve superior alignment between text and imagery.

\subsubsection{De-biasing Results}\label{sec:debiasing}

\Cref{tab:debias_perf_sd15} shows the de-biasing results across eight different professions when using SD1.5.
The values represent the ratio of the minor attribute, \textit{i.e}, for the first four professions where males are dominant, the female ratio is shown.
The bottom row (Avg. $\Delta$) reports the average absolute differences between the ratio and the target ratio of 0.5 across the professions. A lower Avg. value indicates a more balanced ratio between males and females.

The two leftmost columns show that SD tends to amplify biases present in the training data distribution, leading to skewed outcomes.
Compared to the results of the vanilla SD, our results, shown in the rightmost column, show the increased ratio of the minor attribute even without additional training.
The superiority of our method is particularly evident in female-dominant professions (bottom four rows). While training-based de-biasing methods (UCE, FTDiff, and SelfDisc) increase the minor attribute ratios, they rarely reach the target ratio of 0.5, even with additional training. 
FairDiffusion (FairDiff) achieves the same average results as our method but struggles to maintain image-text alignment, as highlighted in \Cref{sec:alignment}.
Our method, on the other hand, achieves both effective de-biasing and robust image generation functionality.
\Cref{tab:debias_perf_general} compares the Avg. $\Delta$ value of vanilla SD and our method, supporting the generalizability of our method across different versions of SD.

\Cref{fig:compare_example} (b) shows the examples generated for two professions.
The images generated from FairDiffusion (middle) present physical characteristics of both women and men simultaneously, such as breasts and mustaches. 
In contrast, our method generates images with clearer gender cues. 
\section{Conclusion}
\label{sec:conclusion}

In this paper, we uncover the built-in potential in SD for training-free de-biasing. 
Based on the identification of minority regions through the proposed mode test, our de-biasing approach efficiently and effectively reduces bias by applying weak guidance to random noise so that it is guided toward the minority regions. 
Experimental results demonstrate that our method consistently reduces bias across multiple versions of Stable Diffusion, underscoring its applicability and robustness.
Additionally, by not requiring additional training, the core image generation functionalities of SD are not compromised.

There are more diverse and complex types of bias, but most de-biasing research focuses on gender and race due to their significant societal impact. 
Although our work follows previous studies by focusing only on gender and race, we believe it can serve as a foundation for future de-biasing research on other biases.

\paragraph{Acknowledgement} Sincere gratitude to Robin Rombach for invaluable discussions. This work was supported by Stability.ai, Institute of Information \& communications Technology Planning \& Evaluation (IITP) grant funded by the Korea government (MSIT) [NO.RS-2021-II211343, Artificial Intelligence Graduate School Program (Seoul National University)], the BK21 FOUR program of the Education and Research Program for Future ICT Pioneers, Seoul National University in 2024, the National Research Foundation of Korea (NRF) grant funded by the Korea government (MSIT) (No. 2022R1A3B1077720, 2022R1A5A708390811), and Samsung Electronics Co., Ltd (IO240124-08661-01).

{
    \small
    \bibliographystyle{ieeenat_fullname}
    \bibliography{main}
}

\clearpage
\setcounter{section}{0}
\renewcommand\thesection{\Alph{section}}
\setcounter{table}{0}
\renewcommand{\thetable}{S\arabic{table}}
\setcounter{figure}{0}
\renewcommand{\thefigure}{S\arabic{figure}}
\maketitlesupplementary

\section{Experimental Details}\label{sec:app_details}
\subsection{Common Settings}
For all experiments conducted with SD1.5, SD2, and SDXL, images are generated using 50 steps with the PNDM scheduler, while SD3 employs the FlowMatchEulerDiscrete scheduler with 28 steps. Image resolutions are set to $512 \times 512$ for SD1.5 and SD2\footnote{\url{https://huggingface.co/stabilityai/stable-diffusion-2-base}}, and $1024 \times 1024$ for SDXL\footnote{\url{https://huggingface.co/stabilityai/stable-diffusion-xl-base-1.0}} and SD3\footnote{\url{https://huggingface.co/stabilityai/stable-diffusion-3-medium-diffusers}}. Unless stated otherwise, we use a CFG scale ($\alpha$) of 6 for SD1.5 and SD2, and 4 for SDXL and SD3. The experiments are performed using NVIDIA RTX 8000, A40, and H100 GPUs.

\subsection{Settings for Comparison Methods}\label{sec:app_expdetail_comparisonmethods}
All the baseline methods we evaluate were originally developed using SD1.4 and SD1.5, so our primary comparisons are conducted using SD1.5. Additionally, we assess our method against FairDiffusion, the only training-free baseline, using SD2 with the same hyperparameters as in SD1.5. Since SDXL features a distinct architecture from SD1.5, we report results only for vanilla SD and our method in this case.

For FairDiffusion, we adopt the default hyperparameters provided in its official implementation. For FTDiff, the pre-trained checkpoints for both the text encoder and LoRA are used. For UCE and SelfDisc, we strictly follow the training and inference procedures outlined in their official codebases. Detailed settings for each method are provided below.

\paragraph{FairDiffusion}\footnote{\url{https://github.com/ml-research/Fair-Diffusion}}
The editing prompts are set to [``\texttt{female person}", ``\texttt{male person}"]. The direction of guidance—subtracting or adding—is randomly selected between the two concepts. The guidance scale for editing is set to 3. The threshold, momentum scale, and momentum beta are set to 0.9, 0.5, and 0.6, respectively. The weights of the individual concepts are set equally to 1.

\paragraph{UCE}\footnote{\url{https://github.com/rohitgandikota/unified-concept-editing}}
We trained UCE from scratch following the instructions provided in the official GitHub repo. 
UCE training requires a pre-defined set of professions and the model reported in the main paper is trained on 36 professions, but for fair comparison with other methods, we remove the professions overlapping with the eight professions used in our paper.

\paragraph{FTDiff}\footnote{\url{https://github.com/sail-sg/finetune-fair-diffusion}}
We use the pre-trained checkpoint provided by the authors. 
FTDiff provides various checkpoints where different components of SD are trained, such as prefix tuning or LoRA fine-tuning.
We used `from-paper\_finetune-text-encoder\_09190215\/checkpoint-9800\_exported\/text\_encoder\_lora\_EMA.pth'.

\paragraph{SelfDisc}\footnote{\url{https://github.com/hangligit/InterpretDiffusion}}
Since the official code does not provide a pre-trained checkpoint, we trained SelfDisc from scratch with default settings and hyperparameters provided by the authors.
Then we generated images by uniformly sampling attributes.

\subsection{Detailed Setting of Our Method}
Our method requires no additional training, making it adaptable to any version of Stable Diffusion with minimal effort. The only parameter to configure is $\tau$, which determines the initial time steps during which $\hat{c}$ is used exclusively. As described in the main text, subsequent time steps alternate between using $\hat{c}$ and $c$, ensuring a balanced integration of guidance throughout the process.

For SD1.5 and SDXL, we set $\tau$ to 0.9. In SD2 and SD3, where the guidance strength for adding attribute directions is weaker, $\tau$ is adjusted to 0 and 0.5, respectively. In the case of SD3, which incorporates three text encoders, we add the attribute direction across all positions for the T5 text encoder. For models like SDXL and SD3, where text encoding involves pooling as part of the conditioning, the attribute direction is applied to the pooled representation.

To maintain the scale of the original text embedding, we renormalize each position so that the norm of $\boldsymbol{c} + \boldsymbol{a}a_k$ matches the norm of $\boldsymbol{c}$. This ensures consistency in the embedding space while incorporating attribute-specific adjustments.

\section{Additional Results for Exploring Guidance to the Minority Region}\label{sec:app_additional_analyses}
\subsection{Noisy Text Condition}\label{sec:app_noisy_text}
\begin{figure*}[t]
  \centering
  \includegraphics[width=0.9\linewidth]{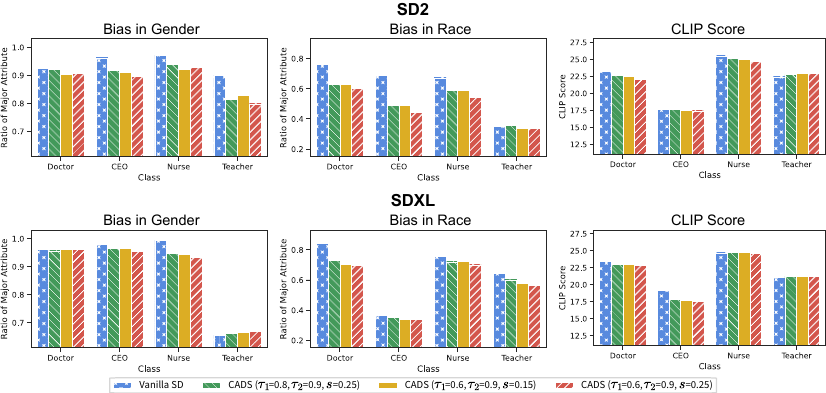}
  \caption{Change in the ratio of major attributes and CLIP score when applying CADS to SD2 and SDXL.}
  \label{fig:cads_app}
\end{figure*}

\Cref{fig:cads_app} demonstrates that introducing noise through CADS effectively reduces gender and racial biases in image generation for both SD2 and SDXL. As discussed in Section 3.2.2 regarding results for SD1.5 and SDXL, increasing the noise level in the text condition (achieved with larger $s$ and smaller $\tau_1$) decreases the prevalence of major attributes, leading to reduced bias in terms of both gender and race. For results using a teacher model, the impact is minimal in certain cases (\textit{e.g.}, racial bias in SD2) or even counterproductive (\textit{e.g.}, increased gender bias in SDXL), particularly when the initial bias in vanilla SD-generated images is less pronounced for specific professions.

Examples of images generated by vanilla SD and CADS, shown in \Cref{fig:cads_example_app}, highlight that CADS produces more diverse outputs and helps mitigate bias. However, CADS occasionally struggles to align the generated images with the provided text prompts, as reflected in a decrease in CLIP scores in \Cref{fig:cads_app}.

These findings indicate that perturbing text conditions with noise, as demonstrated by CADS, can be an effective strategy for reducing bias across different versions of SD. However, as noted in the main text, this approach may come at the cost of weaker alignment between generated images and their text prompts.

\section{Addtional Experimental Results with De-biasing}

\subsection{Image-Text Alignment}\label{sec:app_versatility}
To evaluate the versatility of our method, we generate 100 images with text prompts where the gender is explicitly specified. The image generation process is repeated three times, and the mean and standard deviation of the attribute ratio are reported in \Cref{tab:explicit_attr1,tab:explicit_attr2}.
\Cref{tab:explicit_attr1} provides detailed values illustrated in Figure 8, showing the results for eight professions using SD1.5 with our method and other baselines.
\Cref{tab:explicit_attr2} presents results for SD2, SDXL, and SD3. The results show that our method successfully generates the explicitly specified attribute in the text prompts across all models with a ratio close to 1.0, ensuring versatility. Other methods, including FairDiffusion, fail to achieve a ratio of 1.0, indicating they often fail to generate attributes specified in the text prompts.

\subsection{Debiasing Results}
\Cref{tab:debias_perf_sd2sdxl} shows the de-biasing results on eight different professions using SD2 and SDXL. The numbers indicate the proportion of the minor attribute. It is worth noting that SDXL produces a higher ratio of male teachers compared to female teachers, which differs from SD1.5 and SD2. However, we maintain consistency by reporting the male ratio for teachers. The table demonstrates that our method effectively mitigates bias across different models.

A de-biasing technique should generalize to any prompt susceptible to a stereotype, \textit{e.g.}, it should be able to successfully mitigate bias in images generated from various prompts such as ``\texttt{a photo of a lawyer in a workplace}" and ``\texttt{a portrait of a fashion designer}". 
We test the generalizability with 35 different professions for 5 different prompt templates (``\texttt{a painting of a/an \{profession\}}", ``\texttt{a/an \{profession\} working}", ``\texttt{a/an \{profession\} laughing}", ``\texttt{a/an \{profession\} in the workplace}", and ``\texttt{a/an \{profession\} digital art}"), following \cite{orgad2023editing}. 
For each template, we generate 24 images resulting in a total of 120 images per profession. In \Cref{tab:time}, we present the mean and standard deviation of the female ratio across five prompt templates.

\subsection{Debiasing on Racial Bias}
We additionally evaluate our method within racial bias. Here, we use a name list of attributes: [``\texttt{White person}", ``\texttt{Black person}", ``\texttt{Asian}", ``\texttt{Indian}", ``\texttt{Latino}"]. Other settings are set same as evaluating within gender bias.
\Cref{tab:debias_perf_sd15_race} shows that our method shows reduced ratio of major attribute compared to vanilla SD in all professions.

\subsection{Debiasing Multiple Biases}
We evaluate our method’s ability to debias both gender and race at the same time by sampling two separate weak text embeddings—one for gender attributes and another for race attributes—and then adding them to the original text embedding.

We measure performance by computing the discrepancy score, $\mathcal{D}$, the difference between the generated and target ratio. Formally, $D$ is defined as follows:
\begin{equation}\label{eq:discrepancy}
\mathcal{D}=\frac{1}{|\mathcal{S}|}\sum_{s \in \mathcal{S}}\left\vert{\mathbb{E}_{\mathcal{Y}}\bigl[\mathbbm{1}_{f(y)=s}\bigr]} - \frac{1}{|\mathcal{S}|}\right\vert,
\end{equation}
where $\mathcal{S}$ is a set of attributes, $f(y)$ is the classified attribute of image $y$, and $\mathcal{Y}$ is a set of generated images. A lower $D$ indicates more balanced generation and reduced bias.

\Cref{tab:debias_simultaneous} reports the average discrepancy score over eight professions, which we mainly use to evaluate our method, demonstrating that our method effectively addresses multiple biases at once.

\subsection{Human Evaluation}
We additionally conduct a human evaluation to assess two key aspects: 1) fairness and 2) image diversity and quality. A total of 40 participants took part in the survey. The following paragraphs provide a detailed overview of the survey process and the results.

\begin{figure*}[t]
\centering
\includegraphics[width=0.85\linewidth]{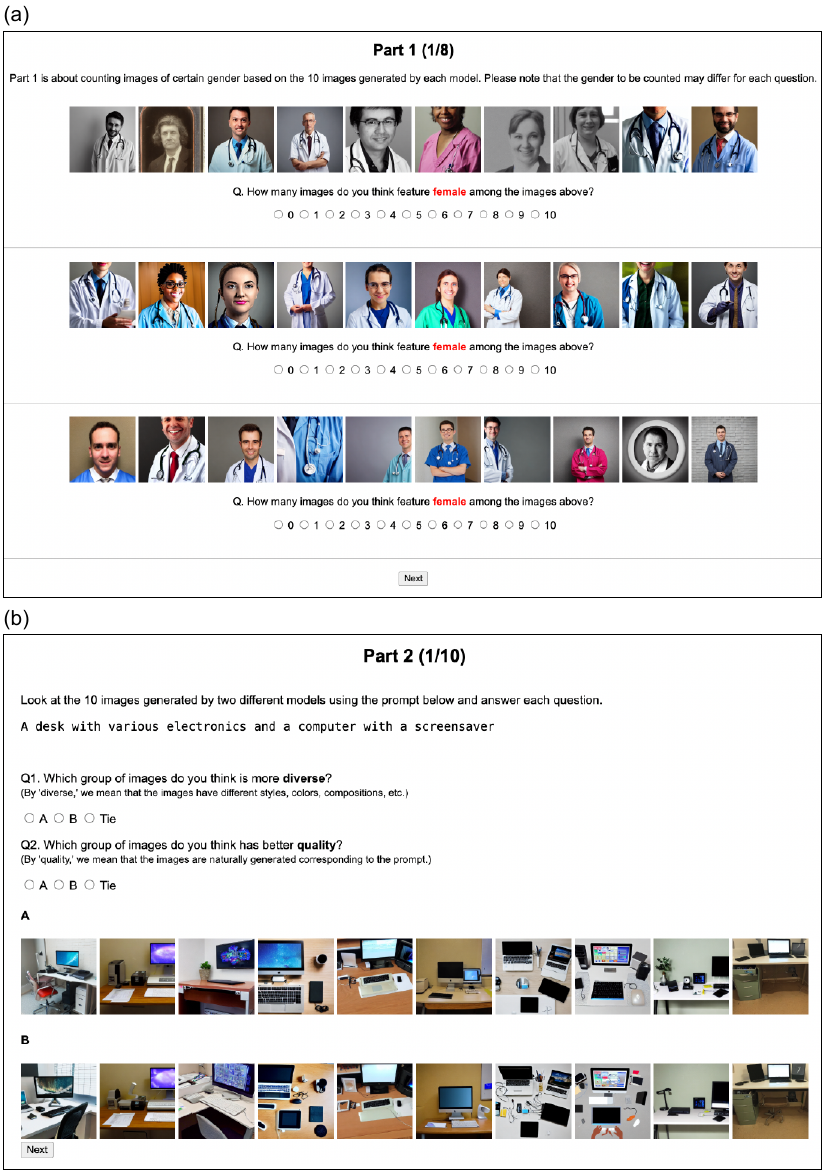}
\caption{Example of a survey page for human evaluation focusing on (a) fairness and (b) image diversity and quality.}
  \label{fig:survey}
\end{figure*}

\begin{figure*}
  \centering
  \includegraphics[width=\linewidth]{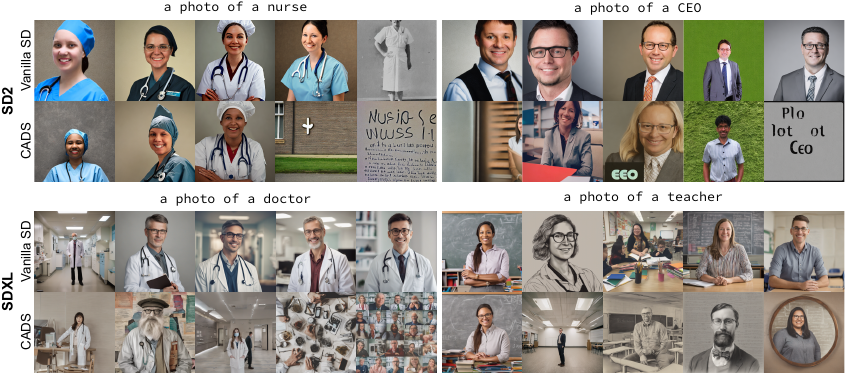}
  \caption{Examples of images generated using vanilla SD and CADS with SD2 and SDXL.}
  \label{fig:cads_example_app}
\end{figure*}

\paragraph{Fairness}
As shown in \Cref{fig:survey}(a), participants were shown 10 images per profession and asked to count the number of images exhibiting minor attributes (gender). The images were generated using one of three methods: vanilla SD, the main competitor FairDiff, and our method. After viewing the images, each participant was asked to count how many images displayed the minor attributes. \Cref{tab:humaneval_fairness} shows the average discrepancy scores over for this task. These human-generated results are consistent with the findings from the attribute classifier, further supporting the effectiveness of our method in reducing bias from a human perspective.

\paragraph{Image Quality and Diversity} 
For evaluating image quality and diversity, we used captions from the COCO-30k dataset. Participants were shown two groups of 10 images each, one generated using our method and the other using vanilla SD. As shown in As shown in \Cref{fig:survey}(b), they were then asked to select the better group based on quality, including text-image alignment, and diversity, with the option to select a tie. As shown in \Cref{tab:humaneval_diversity_quality}, the most common response was ``Tie" across all the results, indicating that our method successfully maintains the image quality and diversity comparable to vanilla SD, as perceived by humans.

\subsection{Qualitative Results}
\Cref{fig:ours_example_app} shows examples of images generated with vanilla SD and our method using SD1.5, SD2, SDXL, and SD3.
The examples demonstrate our method successfully generates images with minor attributes, regardless of the model version used. Images with major attributes are also produced without compromising quality.

\begin{figure*}
  \centering
  \includegraphics[width=0.6\linewidth]{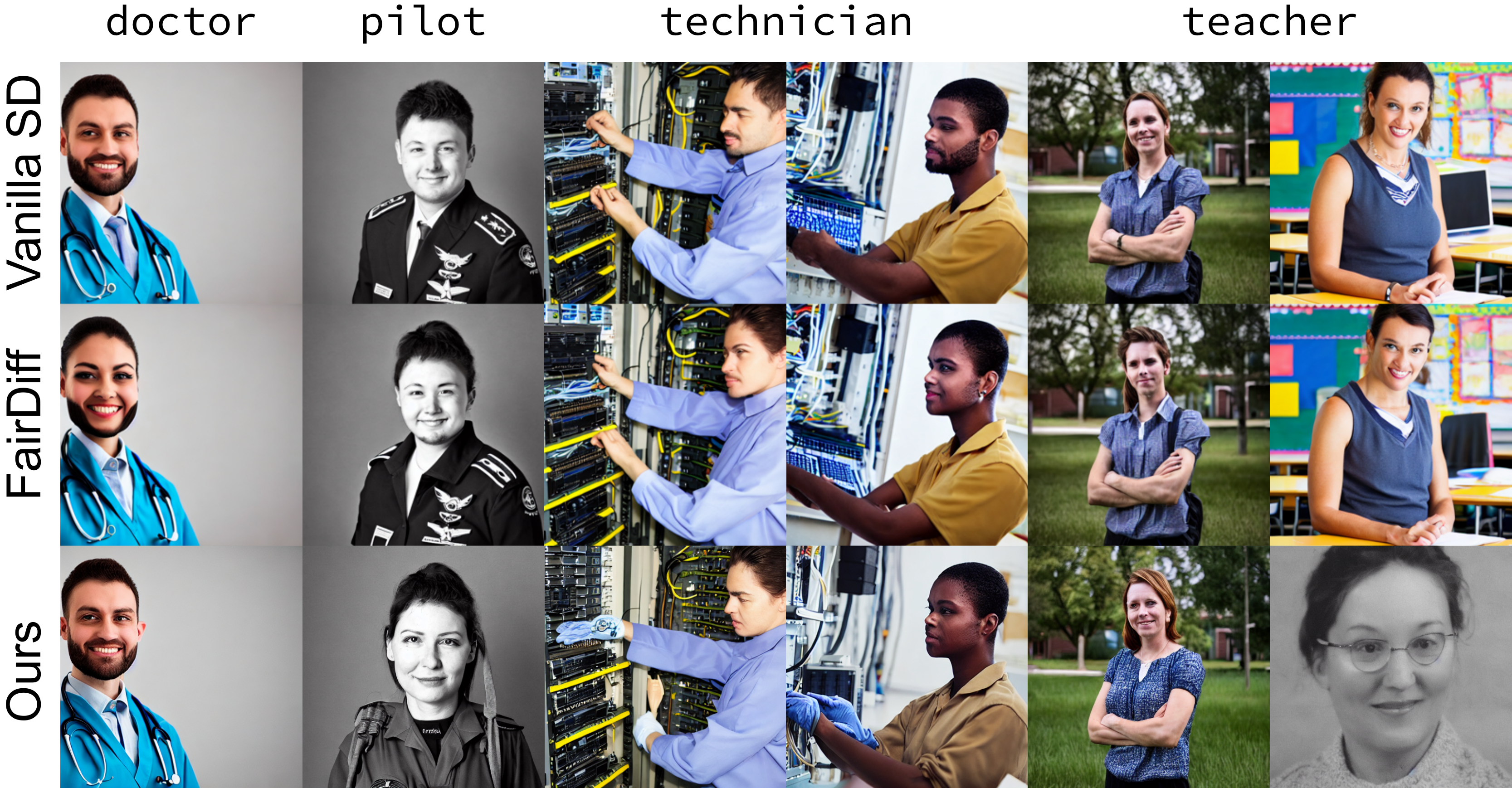}
  \caption{Examples of generated images with vanilla SD, FairDiff, and our method, using SD1.5.}
  \label{fig:compare_example_app}
\end{figure*}

\begin{figure*}
  \centering
  \includegraphics[width=\linewidth]{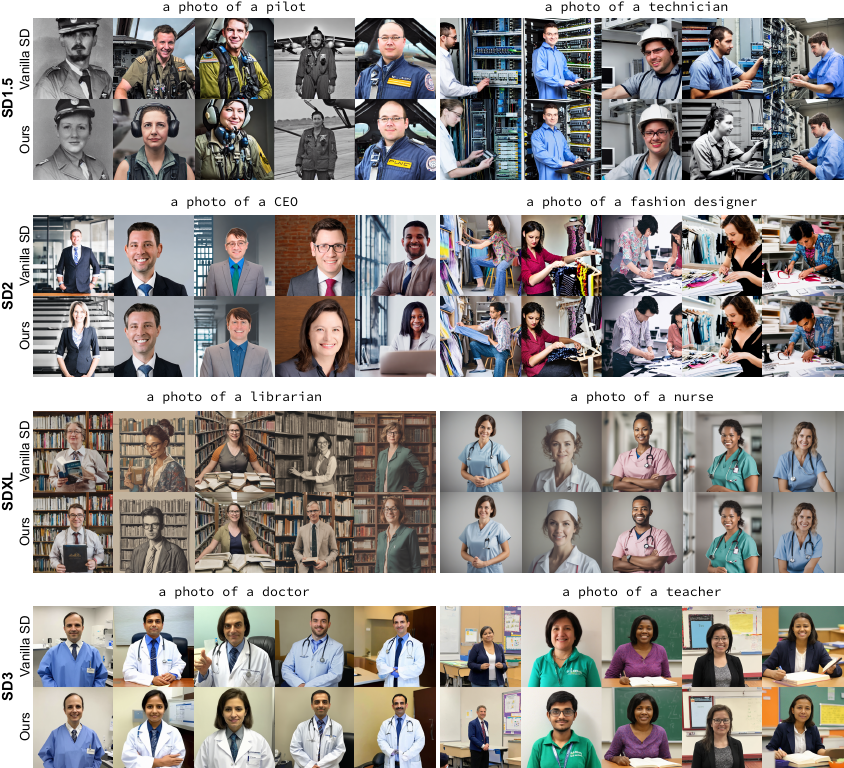}
  \caption{Examples of generated images with vanilla SD and our method.}
  \label{fig:ours_example_app}
\end{figure*}

\paragraph{Qualtitative Comparison with FairDiffusion.}
\Cref{fig:compare_example_app} provide additional qualitative comparisons between FairDiffusion and our method. The individuals in the images generated with FairDiffusion often exhibit physical traits of both women and men simultaneously, while those in the images generated with our method do not.

\newpage
\begin{table}[ht]
  \centering
\begin{tabular}{lcccc}\toprule
$\mathcal{D} (\downarrow)$ &Vanilla SD &Ours-R & Ours-R &Ours-S \\\midrule
Gender &0.404 &0.163 & - &0.157\\
Race  &0.177  & - &0.090  &0.078\\
\bottomrule
  \end{tabular}
    \caption{Average of difference of ratio with target within 1,000 generated images with SD1.5.  ``Ours-\textbf{R}'' and ``Ours-\textbf{S}'' indicate debiasing gender and race \textbf{R}espectively and \textbf{S}imulataneously.}
  \label{tab:debias_simultaneous}
\end{table}

\begin{table}[ht]
  \centering
\begin{tabular}{lccc}
        \toprule
        Fairness &SD &Fairdiff &Ours \\
        \midrule
            $\mathcal{D(\downarrow)}$ &0.341&0.153&0.155\\
        \bottomrule
        \end{tabular}
    \caption{Average discrepancy score across 8 professions from human evaluation, based on 1,000 generated images using SD1.5.}
  \label{tab:humaneval_fairness}
\end{table}

\begin{table}[ht]
  \centering
\begin{tabular}{lccc}
            \toprule
            Ratio &SD &Ours &Tie \\
            \midrule
            Quality &0.30 &0.25 &0.45 \\
            Diversity &0.20 &0.17 &0.63 \\
            \bottomrule
        \end{tabular}
    \caption{Winning ratio, including ties, for better diversity and quality from human evaluation, based on 1,000 generated images using SD1.5.}
  \label{tab:humaneval_diversity_quality}
\end{table}

\begin{table*}
  \centering
\adjustbox{max width=\textwidth}{
\begin{tabular}{c|cccccc}\toprule
Profession &SD &FairDiffusion &FTDiff &SelfDisc &UCE &Ours \\\midrule
\multicolumn{7}{c}{\texttt{A photo of a female \{profession\}}} \\
CEO &1.00±0.01 &0.95±0.02 &0.97±0.00 &1.00±0.00 &0.99±0.00 &0.99±0.01 \\
Doctor &1.00±0.01 &0.89±0.01 &1.00±0.00 &1.00±0.00 &0.99±0.01 &0.99±0.01 \\
Pilot &0.99±0.01 &0.87±0.03 &0.99±0.01 &1.00±0.00 &0.97±0.01 &0.99±0.00 \\
Technician &1.00±0.00 &0.85±0.01 &1.00±0.00 &0.99±0.00 &0.99±0.00 &1.00±0.00 \\
Fashion designer &1.00±0.00 &0.91±0.01 &0.92±0.01 &1.00±0.00 &0.99±0.00 &0.98±0.01 \\
Nurse &1.00±0.00 &0.90±0.01 &0.99±0.00 &1.00±0.00 &0.99±0.01 &0.99±0.01 \\
Librarian &1.00±0.00 &0.93±0.01 &1.00±0.00 &0.99±0.00 &1.00±0.00 &0.99±0.01 \\
Teacher &1.00±0.01 &0.97±0.02 &0.99±0.01 &1.00±0.00 &1.00±0.00 &1.00±0.00 \\\midrule

\multicolumn{7}{c}{\texttt{A photo of a male \{profession\}}}\\
CEO &1.00±0.00 &0.64±0.04 &1.00±0.00 &0.64±0.03 &0.89±0.01 &0.95±0.01 \\
Doctor &1.00±0.00 &0.46±0.06 &0.98±0.01 &0.72±0.03 &0.99±0.01 &0.99±0.01 \\
Pilot &0.99±0.01 &0.29±0.04 &0.89±0.01 &0.51±0.04 &0.94±0.01 &0.91±0.01 \\
Technician &1.00±0.00 &0.39±0.02 &0.98±0.02 &0.67±0.05 &0.99±0.00 &1.00±0.00 \\
Fashion designer &0.99±0.00 &0.54±0.10 &0.94±0.02 &0.66±0.02 &0.74±0.04 &0.98±0.01 \\
Nurse &1.00±0.01 &0.40±0.03 &0.93±0.01 &0.58±0.03 &0.54±0.02 &0.90±0.01 \\
Librarian &1.00±0.00 &0.55±0.04 &1.00±0.00 &0.74±0.02 &0.91±0.04 &1.00±0.00 \\
Teacher &0.96±0.01 &0.50±0.06 &0.86±0.03 &0.64±0.02 &0.98±0.01 &0.99±0.01 \\

\bottomrule
  \end{tabular}
}
  \caption{Ratio of the attributes within images generated by SD1.5 using attribute-specified text prompts. The numerical values represent the attributes specified by the text prompts. Mean and standard deviation of three runs (ratio of each run is obtained using 100 images) are reported.}
  \label{tab:explicit_attr1}
\end{table*}

\begin{table*}
  \centering
\adjustbox{max width=\textwidth}{
\begin{tabular}{c|ccc|ccc}\toprule
\multirow{2}{*}{Profession}& \multicolumn{3}{c}{\texttt{A photo of a female \{profession\}}} & \multicolumn{3}{c}{\texttt{A photo of a male \{profession\}}} \\ 
 &Vanilla SD &FairDiffusion &Ours &Vanilla SD &FairDiffusion &Ours \\\midrule
 \multicolumn{7}{c}{\textbf{SD2}} \\
CEO &1.00±0.01 &0.93±0.02 &0.99±0.00 &1.00±0.00 &0.87±0.05 &1.00±0.00 \\
Doctor &0.99±0.00 &0.57±0.03 &1.00±0.00 &1.00±0.00 &0.87±0.01 &1.00±0.00 \\
Pilot &0.99±0.01 &0.79±0.05 &0.99±0.00 &0.99±0.01 &0.56±0.01 &0.99±0.00 \\
Technician &0.97±0.01 &0.58±0.06 &0.98±0.00 &1.00±0.00 &0.78±0.03 &1.00±0.00 \\
Fashion designer &1.00±0.00 &0.73±0.01 &1.00±0.00 &1.00±0.00 &0.74±0.01 &1.00±0.00 \\
Nurse &0.99±0.01 &0.64±0.03 &1.00±0.00 &1.00±0.00 &0.74±0.04 &1.00±0.00 \\
Librarian &0.99±0.01 &0.79±0.03 &0.99±0.00 &1.00±0.00 &0.78±0.02 &1.00±0.00 \\
Teacher &0.99±0.01 &0.79±0.03 &1.00±0.00 &1.00±0.00 &0.78±0.02 &1.00±0.00 \\
\midrule
\multicolumn{7}{c}{\textbf{SDXL}} \\
CEO &1.00±0.00 &- &1.00±0.00 &1.00±0.00 &- &1.00±0.00 \\
Doctor &0.99±0.01 &- &1.00±0.00 &1.00±0.00 &- &1.00±0.00 \\
Pilot &0.99±0.01 &- &1.00±0.00 &0.99±0.01 &- &0.99±0.01 \\
Technician &1.00±0.00 &- &1.00±0.00 &1.00±0.00 &- &1.00±0.00 \\
Fashion designer &1.00±0.00 &- &1.00±0.00 &1.00±0.00 &- &1.00±0.00 \\
Nurse &1.00±0.00 &- &1.00±0.00 &1.00±0.00 &- &1.00±0.00 \\
Librarian &1.00±0.01 &- &1.00±0.00 &1.00±0.00 &- &1.00±0.00 \\
Teacher &1.00±0.00 &- &1.00±0.00 &1.00±0.00 &- &1.00±0.00 \\
\midrule
\multicolumn{7}{c}{\textbf{SD3}} \\
CEO &1.00±0.00 & - &1.00±0.00 &1.00±0.00 & - &1.00±0.00 \\
Doctor &1.00±0.00 & - &1.00±0.00 &1.00±0.00 & - &1.00±0.00 \\
Pilot &1.00±0.00 & - &1.00±0.00 &1.00±0.00 & - &1.00±0.00 \\
Technician &1.00±0.00 & - &1.00±0.00 &1.00±0.00 & - &1.00±0.00 \\
Fashion designer &1.00±0.00 & - &1.00±0.00 &1.00±0.00 & - &1.00±0.00 \\
Nurse &1.00±0.00 & - &1.00±0.00 &1.00±0.00 & - &1.00±0.00 \\
Librarian &1.00±0.00 & - &1.00±0.00 &1.00±0.00 & - &1.00±0.00 \\
Teacher &1.00±0.00 & - &1.00±0.00 &1.00±0.00 & - &1.00±0.00 \\
\bottomrule
  \end{tabular}
}
  \caption{Ratio of the attributes within images generated by SD2, SDXL, and SD3 using attribute-specified text prompts. The numerical values represent the attributes specified by the text prompts.}
  \label{tab:explicit_attr2}
\end{table*}

\begin{table*}
  \centering
  \begin{tabular}{c|ccc|cc|cc}
    \toprule
    \multirow{2}{*}{Profession} & \multicolumn{3}{c|}{SD2} & \multicolumn{2}{c|}{SDXL} & \multicolumn{2}{c}{SD3}\\
    & Vanilla SD & FairDiffusion & Ours & Vanilla SD & Ours & Vanilla SD & Ours\\
    \midrule
CEO &0.038 &0.176 &0.147 &0.020 &0.201 &0.000 &0.262 \\
Doctor &0.075 &0.299 &0.258 &0.039 &0.250 &0.126 &0.286 \\
Pilot &0.065 &0.504 &0.159 &0.064 &0.286 &0.017 &0.295 \\
Technician &0.007 &0.333 &0.101 &0.002 &0.011 &0.000 &0.029 \\
Fashion designer &0.062 &0.088 &0.152 &0.270 &0.384 &0.002 &0.008 \\
Librarian &0.071 &0.387 &0.142 &0.345 &0.448 &0.000 &0.097 \\
Teacher &0.100 &0.486 &0.128 &0.653 &0.541 &0.008 &0.228 \\
Nurse &0.029 &0.338 &0.021 &0.006 &0.025 &0.000 &0.001 \\
    \bottomrule
  \end{tabular}
  \caption{Ratio of minor attribute within 1000 generated images with SD2, SDXL, and SD3.}
  \label{tab:debias_perf_sd2sdxl}
\end{table*}

\begin{table*}
  \centering
\adjustbox{max width=\textwidth}{
\begin{tabular}{c|cccccc}\toprule
Profession&Vanilla SD &FairDiffusion &UCE &FTDiff &SelfDisc &Ours \\\midrule
Attendant &0.10$\pm$0.04 &0.48$\pm$0.09 &0.49$\pm$0.05 &0.25$\pm$0.12 &0.39$\pm$0.19 &0.13$\pm$0.09 \\
Cashier &0.61$\pm$0.21 &0.65$\pm$0.07 &0.49$\pm$0.27 &0.50$\pm$0.07 &0.16$\pm$0.15 &0.66$\pm$0.26 \\
Teacher &0.67$\pm$0.25 &0.69$\pm$0.14 &0.49$\pm$0.32 &0.67$\pm$0.12 &0.10$\pm$0.09 &0.63$\pm$0.32 \\
Nurse &0.98$\pm$0.02 &0.84$\pm$0.07 &0.03$\pm$0.05 &0.81$\pm$0.10 &0.82$\pm$0.18 &0.98$\pm$0.02 \\
Assistant &0.45$\pm$0.18 &0.63$\pm$0.03 &0.49$\pm$0.36 &0.42$\pm$0.15 &0.04$\pm$0.07 &0.46$\pm$0.14 \\
Secretary &0.93$\pm$0.15 &0.86$\pm$0.06 &0.08$\pm$0.07 &0.60$\pm$0.17 &0.16$\pm$0.11 &0.91$\pm$0.07 \\
Cleaner &0.34$\pm$0.18 &0.45$\pm$0.13 &0.68$\pm$0.18 &0.47$\pm$0.22 &0.46$\pm$0.11 &0.25$\pm$0.16 \\
Receptionist &0.93$\pm$0.07 &0.86$\pm$0.09 &0.33$\pm$0.23 &0.73$\pm$0.18 &0.02$\pm$0.04 &0.97$\pm$0.05 \\
Clerk &0.40$\pm$0.25 &0.61$\pm$0.12 &0.31$\pm$0.16 &0.43$\pm$0.09 &0.63$\pm$0.06 &0.38$\pm$0.19 \\
Counselor &0.48$\pm$0.14 &0.54$\pm$0.05 &0.34$\pm$0.21 &0.46$\pm$0.17 &0.37$\pm$0.19 &0.50$\pm$0.15 \\
Designer &0.43$\pm$0.32 &0.65$\pm$0.13 &0.42$\pm$0.32 &0.53$\pm$0.22 &0.03$\pm$0.04 &0.52$\pm$0.37 \\
Hairdresser &0.64$\pm$0.21 &0.55$\pm$0.17 &0.36$\pm$0.17 &0.63$\pm$0.26 &0.34$\pm$0.08 &0.63$\pm$0.21 \\
Writer &0.39$\pm$0.12 &0.58$\pm$0.07 &0.55$\pm$0.28 &0.58$\pm$0.23 &0.30$\pm$0.15 &0.44$\pm$0.19 \\
Housekeeper &0.98$\pm$0.04 &0.73$\pm$0.17 &0.07$\pm$0.05 &0.95$\pm$0.03 &0.33$\pm$0.19 &0.96$\pm$0.05 \\
Baker &0.45$\pm$0.19 &0.61$\pm$0.11 &0.36$\pm$0.14 &0.65$\pm$0.14 &0.40$\pm$0.08 &0.37$\pm$0.32 \\
Librarian &0.83$\pm$0.22 &0.80$\pm$0.19 &0.08$\pm$0.04 &0.62$\pm$0.19 &0.75$\pm$0.09 &0.85$\pm$0.15 \\
Tailor &0.24$\pm$0.25 &0.53$\pm$0.22 &0.61$\pm$0.26 &0.33$\pm$0.27 &0.18$\pm$0.10 &0.23$\pm$0.24 \\
Driver &0.06$\pm$0.09 &0.43$\pm$0.09 &0.79$\pm$0.11 &0.30$\pm$0.17 &0.20$\pm$0.07 &0.06$\pm$0.07 \\
Supervisor &0.18$\pm$0.14 &0.49$\pm$0.12 &0.61$\pm$0.18 &0.15$\pm$0.16 &0.63$\pm$0.15 &0.13$\pm$0.10 \\
Janitor &0.03$\pm$0.04 &0.19$\pm$0.09 &0.54$\pm$0.22 &0.07$\pm$0.09 &0.02$\pm$0.02 &0.03$\pm$0.03 \\
Cook &0.32$\pm$0.18 &0.48$\pm$0.19 &0.33$\pm$0.32 &0.52$\pm$0.08 &0.31$\pm$0.16 &0.38$\pm$0.20 \\
Laborer &0.03$\pm$0.04 &0.34$\pm$0.10 &0.82$\pm$0.23 &0.15$\pm$0.17 &0.48$\pm$0.14 &0.03$\pm$0.05 \\
Construction worker &0.00$\pm$0.00 &0.21$\pm$0.16 &0.76$\pm$0.28 &0.05$\pm$0.07 &0.06$\pm$0.06 &0.00$\pm$0.00 \\
Developer &0.14$\pm$0.15 &0.38$\pm$0.05 &0.65$\pm$0.18 &0.25$\pm$0.25 &0.39$\pm$0.22 &0.07$\pm$0.08 \\
Carpenter &0.02$\pm$0.04 &0.39$\pm$0.17 &0.70$\pm$0.07 &0.07$\pm$0.13 &0.44$\pm$0.11 &0.03$\pm$0.05 \\
Manager &0.16$\pm$0.16 &0.45$\pm$0.13 &0.72$\pm$0.22 &0.27$\pm$0.17 &0.53$\pm$0.10 &0.24$\pm$0.26 \\
Lawyer &0.13$\pm$0.14 &0.43$\pm$0.08 &0.81$\pm$0.14 &0.24$\pm$0.08 &0.64$\pm$0.22 &0.22$\pm$0.18 \\
Farmer &0.02$\pm$0.02 &0.32$\pm$0.09 &0.97$\pm$0.03 &0.14$\pm$0.07 &0.41$\pm$0.15 &0.02$\pm$0.02 \\
Salesperson &0.09$\pm$0.07 &0.44$\pm$0.15 &0.68$\pm$0.13 &0.16$\pm$0.06 &0.44$\pm$0.14 &0.13$\pm$0.13 \\
Physician &0.10$\pm$0.11 &0.48$\pm$0.18 &0.76$\pm$0.20 &0.19$\pm$0.07 &0.70$\pm$0.10 &0.15$\pm$0.12 \\
Guard &0.03$\pm$0.02 &0.30$\pm$0.11 &0.84$\pm$0.10 &0.09$\pm$0.05 &0.64$\pm$0.09 &0.03$\pm$0.03 \\
Analyst &0.08$\pm$0.08 &0.49$\pm$0.05 &0.71$\pm$0.22 &0.17$\pm$0.11 &0.42$\pm$0.14 &0.09$\pm$0.08 \\
Mechanic &0.03$\pm$0.03 &0.43$\pm$0.12 &0.67$\pm$0.15 &0.08$\pm$0.10 &0.71$\pm$0.11 &0.03$\pm$0.03 \\
Sheriff &0.03$\pm$0.02 &0.49$\pm$0.09 &0.64$\pm$0.18 &0.08$\pm$0.10 &0.92$\pm$0.04 &0.04$\pm$0.07 \\
CEO &0.03$\pm$0.05 &0.31$\pm$0.13 &0.89$\pm$0.06 &0.13$\pm$0.08 &0.63$\pm$0.17 &0.06$\pm$0.09 \\
\bottomrule
  \end{tabular}
}
  \caption{Ratio of female attribute across five text prompts. Images are generated using SD1.5.}
  \label{tab:time}
\end{table*}
\begin{table*}
  \centering
  \adjustbox{max width=\textwidth}{
  \begin{tabular}{c|ccccc|ccccc}
    \toprule
    \multirow{2}{*}{Profession} &\multicolumn{5}{c}{Vanilla SD} &\multicolumn{5}{c}{Ours}\\
&White &Black &Indian &Asian &Latino &White &Black &Indian &Asian &Latino \\\midrule 
CEO &0.091 &0.165 &0.015 &0.228 &\textbf{0.501} &0.025 &\textbf{0.313} &0.087 &0.330 &0.245 \\
Doctor &\textbf{0.660} &0.049 &0.014 &0.031 &0.246 &0.194 &\textbf{0.268} &0.126 &0.186 &0.226 \\
Pilot &\textbf{0.389} &0.028 &0.111 &0.259 &0.213 &0.110 &0.203 &0.234 &\textbf{0.301} &0.152 \\
Technician &\textbf{0.488} &0.060 &0.004 &0.016 &0.432 &0.172 &0.271 &0.052 &0.177 &\textbf{0.328} \\
Fashion designer &0.222 &0.075 &0.030 &\textbf{0.549} &0.124 &0.095 &0.236 &0.105 &\textbf{0.473} &0.091 \\
Nurse &\textbf{0.716} &0.132 &0.010 &0.099 &0.043 &0.269 &0.282 &0.103 &\textbf{0.289} &0.057 \\
Librarian &\textbf{0.886} &0.010 &0.000 &0.045 &0.059 &\textbf{0.516} &0.173 &0.014 &0.203 &0.094 \\
Teacher &\textbf{0.432} &0.125 &0.005 &0.120 &0.318 &0.128 &0.245 &0.064 &\textbf{0.333} &0.230 \\
    \bottomrule
  \end{tabular}}
  \caption{Ratio of each attribute within 1000 generated images with SD1.5. The bold indicates the highest ratio among attributes.}
  \label{tab:debias_perf_sd15_race}
\end{table*}



\end{document}